\documentclass{article}

\usepackage{microtype}
\usepackage{graphicx}
\usepackage{booktabs}
\usepackage{multirow}
\usepackage{array}
\usepackage{hyperref}

\usepackage[accepted]{icml2026}
\usepackage{amsmath}
\usepackage{amssymb}
\usepackage{mathtools}
\usepackage{algorithm}
\usepackage{algorithmic}
\usepackage{subcaption}

\usepackage{float}
\usepackage{placeins}
\usepackage[capitalize,noabbrev]{cleveref}

\icmltitlerunning{Beyond Forecasting: The Belief-to-Trade Layer in Prediction-Market Agents}

\begin{document}

\twocolumn[
  \icmltitle{Beyond Forecasting: The Belief-to-Trade Layer in Prediction-Market Agents}

  \begin{icmlauthorlist}
    \icmlauthor{Yishu Wang}{hkust}
    \icmlauthor{Yuxuan Wang}{pku}
    \icmlauthor{Jiaqi Deng}{hku}
    \icmlauthor{Hanyang Tang}{mit}
  \end{icmlauthorlist}

  \icmlaffiliation{hkust}{Hong Kong University of Science and Technology}
  \icmlaffiliation{pku}{Peking University}
  \icmlaffiliation{hku}{The University of Hong Kong}
  \icmlaffiliation{mit}{Massachusetts Institute of Technology}

  \icmlcorrespondingauthor{Yishu Wang}{issue.00.gui@gmail.com}
  \icmlcorrespondingauthor{Yuxuan Wang}{yuxwang@pku.edu.cn}
  \icmlcorrespondingauthor{Jiaqi Deng}{jiaqideng@connect.hku.hk}
  \icmlcorrespondingauthor{Hanyang Tang}{hytangs@mit.edu}

  \icmlkeywords{AI forecasting, prediction markets, agentic systems, decision theory, risk management}

  \vskip 0.3in
]

\makeatletter
\renewcommand{\ICML@appearing}{\textit{ICML 2026 Workshop on AI Forecasting},
Seoul, South Korea, 2026.}
\renewcommand{\printAffiliationsAndNotice}[1]{\global\icml@noticeprintedtrue%
  \stepcounter{@affiliationcounter}%
  {\let\thefootnote\relax\footnotetext{\raggedright\hspace*{-\footnotesep}\ificmlshowauthors #1\fi%
      \forloop{@affilnum}{1}{\value{@affilnum} < \value{@affiliationcounter}}{
        \textsuperscript{\arabic{@affilnum}}\ifcsname @affilname\the@affilnum\endcsname%
          \csname @affilname\the@affilnum\endcsname%
        \else
          {\bf AUTHORERR: Missing \textbackslash{}icmlaffiliation.}
        \fi
      }.%
      \ifdefined\icmlcorrespondingauthor@text
         \\Correspondence to: \icmlcorrespondingauthor@text.
      \fi
      \ \\[4pt]
      \Notice@String
    }
  }
}
\gdef\icmlcorrespondingauthor@text{%
\mbox{Yishu Wang \textless{}issue.00.gui@gmail.com\textgreater{}},
\mbox{Yuxuan Wang \textless{}yuxwang@pku.edu.cn\textgreater{}},
\mbox{Jiaqi Deng \textless{}jiaqideng@connect.hku.hk\textgreater{}},
\mbox{Hanyang Tang \textless{}hytangs@mit.edu\textgreater{}}}
\makeatother
\printAffiliationsAndNotice{\raggedright}

\begin{abstract}
Forecasting future events has attracted growing attention as a testbed for general-purpose AI. A natural way to ground this evaluation is let the models trade in the prediction markets. 
Trading, however, requires more than forecasting.
Moreover, recent benchmarks report a substantial gap between calibrated probability scores and the trading results.
We propose Raven-Agent, to the best of our knowledge, the first autonomous trading agent for prediction markets.
On a controlled replay over an archived decision set, our architecture achieves the only positive return and the only positive risk-adjusted return among all tested policies.
We have released our code in \href{https://github.com/Alchemist-X/predict-raven}{https://github.com/Alchemist-X/predict-raven}.
\end{abstract}

\section{Introduction}

Forecasting future events has attracted growing attention as a testbed for general-purpose AI \citep{karger2024forecastbench,yang2025prophet}.
The task demands retrieval, temporal reasoning, and decision under uncertainty.
It is naturally studied as a joint problem of probability estimation and action under that estimate.

To make progress on this problem, one line of exisiting works focuses on probability quality and improves LLM forecasters on resolved events.
The Belief-Logit-Forecaster (BLF) of \citet{murphy2026blf} uses structured belief states and multi-trial aggregation, currently leading ForecastBench \citep{karger2024forecastbench}.
Yet such systems improve calibration without acting on their forecasts.
On the other hand, prediction market benchmarks score systems by realized trading return as well as by probability \citep{yang2025prophet,cheng2026polybench,zhang2026predictionarena}.
They consistently report that strong probability scores do not translate into profitable trades.
However, they treat the trading rule as a fixed protocol such as a per market bet or a confidence threshold, not as a designable object.
Moreover, position sizing, portfolio exposure, and risk control are either absent or expressed as a prompt instruction in these papers.
We refer to these downstream components as the \emph{trading layer}, and we make it the object of study in this paper.

Across both lines, the trading layer is either omitted or hardcoded, never treated as a designable object.
A useful trading layer should be explicit and modular, with selection, sizing, and risk control as separate components.
It should be deterministic, so that risk constraints remain in effect even when the model is confidently wrong.
It should also be composable with any forecaster, so that improvements in calibration and in trade execution can be evaluated independently.

In this paper, we propose Raven-Agent, an autonomous prediction market agent built around an explicit, deterministic, composable trading layer.
Within the layer, a ranking module addresses selection by scoring each candidate by a time normalized return rate.
The same module addresses sizing with a fractional Kelly bet on the retained positions.
An execution and risk module addresses risk control by enforcing deterministic constraints on stake, exposure, stop loss, and drawdown outside the language model prompt.
We also include an information module for evidence retrieval and a forecasting module that produces per contract probabilities.
We validate Raven-Agent on a controlled replay protocol. The results show our method improves return on stake and risk adjusted return over other heuristic policies.
Our contributions can be summarized in three-folds:
\begin{itemize}
  \item We propose a replay protocol that reuses archived forecasts to isolate the trading layer from the forecaster.
  \item We propose Raven-Agent, an autonomous agent that makes the trading layer an explicit module composable with a swappable forecaster.
  \item On a controlled replay, the trading layer improves return on stake and risk adjusted return over other policies.
\end{itemize}

\section{Problem Setup}

\subsection{Prediction Markets}

We consider binary prediction markets in which the selected contract pays one dollar if it resolves YES and zero otherwise.
For each market $i$, we observe the realized outcome of the chosen side, written $z_i \in \{0, 1\}$.
The entry price is $q_i \in (0, 1)$, and we let $p_i \in (0, 1)$ denote the model probability that the contract pays.
A stake of $s_i$ dollars buys $s_i / q_i$ shares.
The realized profit at resolution is
\begin{equation}
  \Pi_i = s_i \left(\frac{z_i}{q_i} - 1\right).
  \label{eq:resolved-profit}
\end{equation}
For markets unresolved at the evaluation date, we substitute the mark to market price $m_i$ for $z_i$ in \cref{eq:resolved-profit}.
Resolved and unresolved positions are reported separately throughout.

To compare candidates whose payoff dates differ widely, we need a return rate, not just an edge.
The per dollar expected return of a long position is $p_i / q_i - 1$ before fees and slippage.
The raw probability edge $e_i = p_i - q_i$ does not by itself reflect that capital is locked until resolution.
A small edge that resolves quickly can dominate a larger edge that ties capital for months.
With $T_i$ days to resolution, we therefore rank candidates by a time normalized score
\begin{equation}
  \rho_i = \left(\frac{p_i}{q_i} - 1\right) \cdot \frac{30}{\max(T_i, 1)}.
  \label{eq:monthly-return}
\end{equation}
\emph{This is the per dollar expected return of a candidate trade in units of one month.}
It is a ranking device, not a claim of exact monthly compounding.

\subsection{Trading policies}

A trading policy formalizes which action follows from a forecast.
We write it as$\pi: (p_i, q_i, T_i, \ell_i, W, \mathcal{S}) \mapsto (a_i, s_i)$,
where $\ell_i$ is liquidity, $W$ is bankroll, and $\mathcal{S}$ is the current portfolio state.
The action $a_i$ is either open or skip, and $s_i \geq 0$ is the stake.
Two agents with the same forecasts can produce very different outcomes if they differ in their policy.
We hold all positions to resolution in the main replay and defer dynamic exit to future work.

\section{Methodology}
\label{sec:method}

In this section, we instantiate the trading agent Raven-Agent.
The full loop is summarized in \cref{alg:loop}, and the four modules are described below.

\begin{algorithm}[t]
\caption{Raven-Agent forecast to trade loop.}
\label{alg:loop}
\begin{algorithmic}[1]
\REQUIRE bankroll $W$ and current portfolio state
\STATE Scan candidate markets and retrieve evidence
\STATE Estimate $p_i$ for each candidate $i$ from the candidate lists.
\STATE Compute $\rho_i$ via \cref{eq:monthly-return}; drop candidates below a threshold and rank the rest by $\rho_i$.
\STATE Size each surviving candidate by a quarter Kelly fraction, clipped by liquidity and a per round batch cap.
\STATE Reject any proposal violating exposure, stop loss, or drawdown constraints.
\STATE Submit the remaining orders and review the portfolio state, checking if we need adjust any positions.
\STATE Repeat as new candidate markets and evidence arrive.
\end{algorithmic}
\end{algorithm}


\subsection{Information collection}
\label{sec:method-information}

The information module scans candidate markets and writes a timestamped pulse.
We periodically query the Polymarket Gamma API for each candidate, retrieving its price, liquidity, fee schedule, and resolution date.
A liquidity filter drops markets below a threshold, and an evidence retriever pulls recent news for the survivors.
This follows the spirit of retrieval augmented forecasting \citep{murphy2026blf}.
The pulse serves as a contract between data sources and downstream modules.
New evidence feeds can therefore be added without changing the trading interface.

\subsection{Probability analysis}
\label{sec:method-forecast}

The forecasting module emits the per contract probability $p_i$ for the side the agent prefers to hold.
It is a swappable layer, and we currently support two runtimes.
One parses the pulse markdown directly and combines model side estimates with market implied priors.
The other dispatches the pulse to an external language model through a command line bridge.
Both record, alongside $p_i$, the chosen side, a confidence level, and a short thesis for later audit.
Forecaster centric systems like BLF \citep{murphy2026blf} improve the probability itself via belief states and multi trial aggregation.
Our contribution operates downstream of the forecaster, so any such forecaster can be substituted without changing the trading layer.
In the present replay we hold this module fixed and reuse archived probabilities, so any difference in outcome is attributable to the trading layer.

\subsection{Ranking and sizing}
\label{sec:method-rank}

The ranking module both selects candidates and sets their stakes.
For selection, the module ranks candidates by $\rho_i$ from \cref{eq:monthly-return}, drops those below a threshold, and keeps the top $K = 4$.
Ranking by $\rho_i$ rather than by the raw edge $e_i = p_i - q_i$ accounts for how long capital is locked.
This favors short resolution markets when edges are similar.

For sizing, the module applies one quarter of the Kelly bet to each retained candidate:
\begin{equation}
  s_i = \frac{1}{4} \cdot \max\!\left(0, \frac{p_i - q_i}{1 - q_i}\right) \cdot W,
  \label{eq:quarter-kelly}
\end{equation}
clipped by available liquidity and a per round batch cap.
The Kelly criterion gives the bet size that maximizes long run log wealth under a known win probability \citep{kelly1956new}.
We use one quarter rather than full Kelly because $p_i$ is itself an estimate.
Scaling down the Kelly bet is the standard practitioner correction for input noise.

\subsection{Execution and risk}
\label{sec:method-risk}

The execution and risk module enforces deterministic risk constraints on every order before submission.
We implement it as a set of services that check each proposed order against five hard constraints.
A per trade notional cap and an aggregate exposure cap limit single stake and total open notional respectively.
A per event cap limits capital across correlated contracts of the same event.
A position level stop loss closes any position whose mark to market loss exceeds a fixed fraction of its cost basis.
A portfolio level drawdown halt freezes new openings when the equity drawdown $(W_{\max} - W)/W_{\max}$ exceeds a threshold.
The module also rejects orders against stale pulses or contracts outside the allowlist, and routes surviving orders as fill or kill market orders.
Prior work shows that prompt-level risk guidance is unreliable under end-to-end LLM control \citep{zhang2026predictionarena}, and that optimizing prediction accuracy can reduce trading return when the objective is misaligned \citep{jang2025losingwinner}.
Keeping these checks outside the prompt avoids that failure mode and produces auditable rejection reasons that a learned policy cannot easily match.

\section{Experiments}
\label{sec:experiments}

\subsection{Setup}

\paragraph{Environment.}
We evaluate on a counterfactual replay of archived live decisions from a Polymarket deployment.
The archive records 59 open decisions across 65 pulse runs; 44 reach the executable stage after filtering by liquidity, resolution clarity, and event overlap.
For each candidate, Raven-Agent estimated the payout probability, compared it with the market price, computed the edge, and proposed a stake; live trades were submitted through Polymarket's CLOB API.
The replay reuses the archived forecasts, prices, and execution records from these runs (concrete examples in \cref{app:archive}).

All policies face the same archived markets, timestamps, prices, and probability estimates; the only difference is the rule used to decide which positive-edge forecasts become trades and how much capital they receive.
Positions are held to resolution where available, and otherwise marked to market at a fixed evaluation date.

\paragraph{Policies.}
We compare five policies.
\textbf{i)} \emph{Forecast-only} trades every positive-edge forecast at a fixed \$10 stake.
\textbf{ii)} \emph{Edge-proportional} trades the same set but sizes each stake proportionally to the probability edge $e_i = p_i - q_i$, with total stake normalized to match Forecast-only (\$580) so that ROI differences reflect allocation, not capital deployed.
\textbf{iii)} \emph{Edge-filter} keeps only forecasts whose edge exceeds five percentage points, at a fixed \$10 stake.
\textbf{iv)} \emph{Raven-Agent (fixed)} applies the full execution and risk layer of Raven-Agent at a fixed \$10 stake per surviving trade.
It rejects forecasts when market data is stale, model confidence is low, liquidity is insufficient, or event exposure is excessive.
\textbf{v)} \emph{Raven-Agent (full)} uses the same execution layer with Raven's deployed quarter-Kelly sizing, so capital scales with edge, time to resolution, liquidity, and portfolio risk.

\paragraph{Metrics.}
We report three metrics.
Probability quality is measured by the Brier score $\mathrm{Brier} = \tfrac{1}{M}\sum_i (p_i - z_i)^2$ on resolved rows \citep{brier1950verification}.
Trading quality is measured by return on stake $\mathrm{ROI} = \sum_i \Pi_i / \sum_i s_i$.
Risk adjusted return is measured by a stake weighted Sharpe ratio $\mathrm{Sharpe}_w = \bar{r}_w / \mathrm{std}_w(r_i)$, where $r_i$ is the realized or mark to market trade return and $w_i = s_i$.
We abbreviate $\mathrm{Sharpe}_w$ as $S_w$ and report the aggregate $\sum_i \Pi_i$ as profit and loss (PnL) in the tables.

\subsection{Result Analysis}
\label{sec:main-results}

\begin{table*}[!t]
  \centering
  \caption{Trading layer ablation on the main archive. Edge-proportional total stake is normalized to \$580 to match Forecast-only.}
  \label{tab:main-results}
  \small
  \begin{tabular}{@{}p{0.30\textwidth}rrrrrr@{}}
    \toprule
    \textbf{Policy} & \textbf{ROI} & \textbf{Selected Brier$^\dagger$} & \textbf{$S_w$} & \textbf{\# Exec.} & \textbf{PnL} & \textbf{Stake} \\
    \midrule
    Forecast-only & $-$10.7\% & 0.329 & $-$0.20 & 58 & $-$\$62 & \$580 \\
    Edge-proportional & $-$55.5\% & 0.329 & $-$0.91 & 58 & $-$\$322 & \$580 \\
    Edge-filter & $-$9.3\% & 0.302 & $-$0.16 & 48 & $-$\$45 & \$480 \\
    Raven-Agent (fixed) & $-$4.7\% & \textbf{0.258} & $-$0.09 & 42 & $-$\$20 & \$420 \\
    Raven-Agent (full) & \textbf{+15.9\%} & \textbf{0.258} & \textbf{+0.42} & 42 & +\$131 & \$820 \\
    \bottomrule
  \end{tabular}
  \par\vspace{4pt}
  {\footnotesize $^\dagger$ Brier score computed on each policy's traded subset of resolved rows. All policies use the same archived probabilities; the full-archive Brier is 0.329 for every policy. See \S\ref{sec:main-results} for discussion.}
\end{table*}

\begin{figure*}[!t]
\centering
\includegraphics[width=\linewidth]{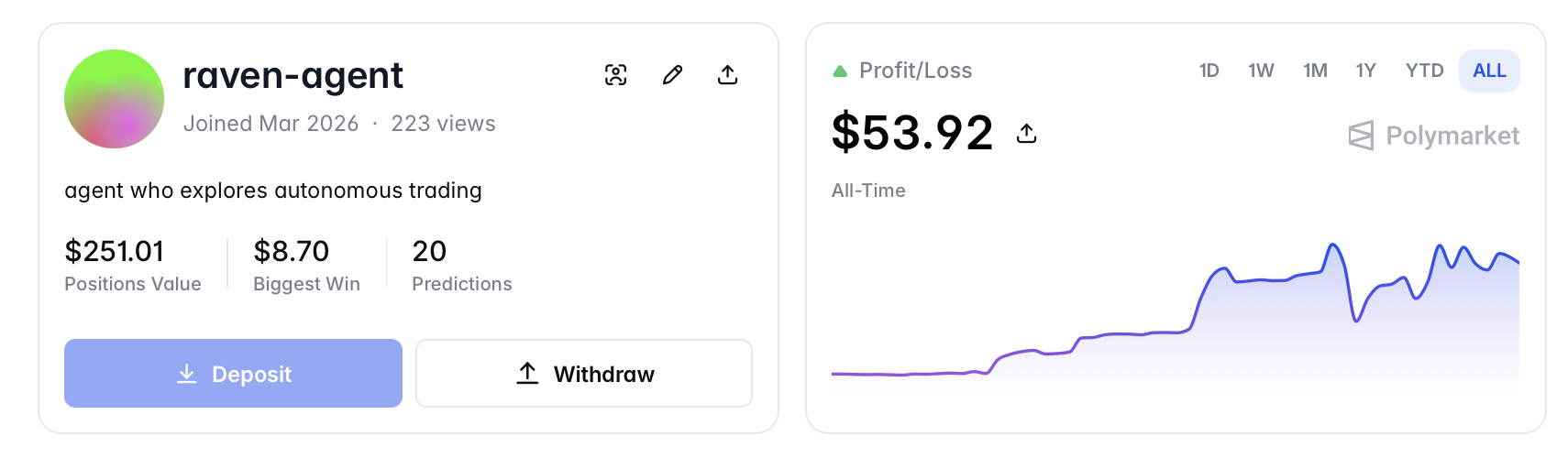}
\caption{Raven-Agent's live Polymarket profile and all-time profit and loss curve. Across 20 predictions, the agent currently holds \$251.01 in open positions and has accumulated \$53.92 in cumulative profit.}
\label{fig:pnl-curve}
\end{figure*}

\begin{figure*}[!t]
  \centering
  \begin{subfigure}[t]{0.48\textwidth}
    \centering
    \includegraphics[width=\linewidth]{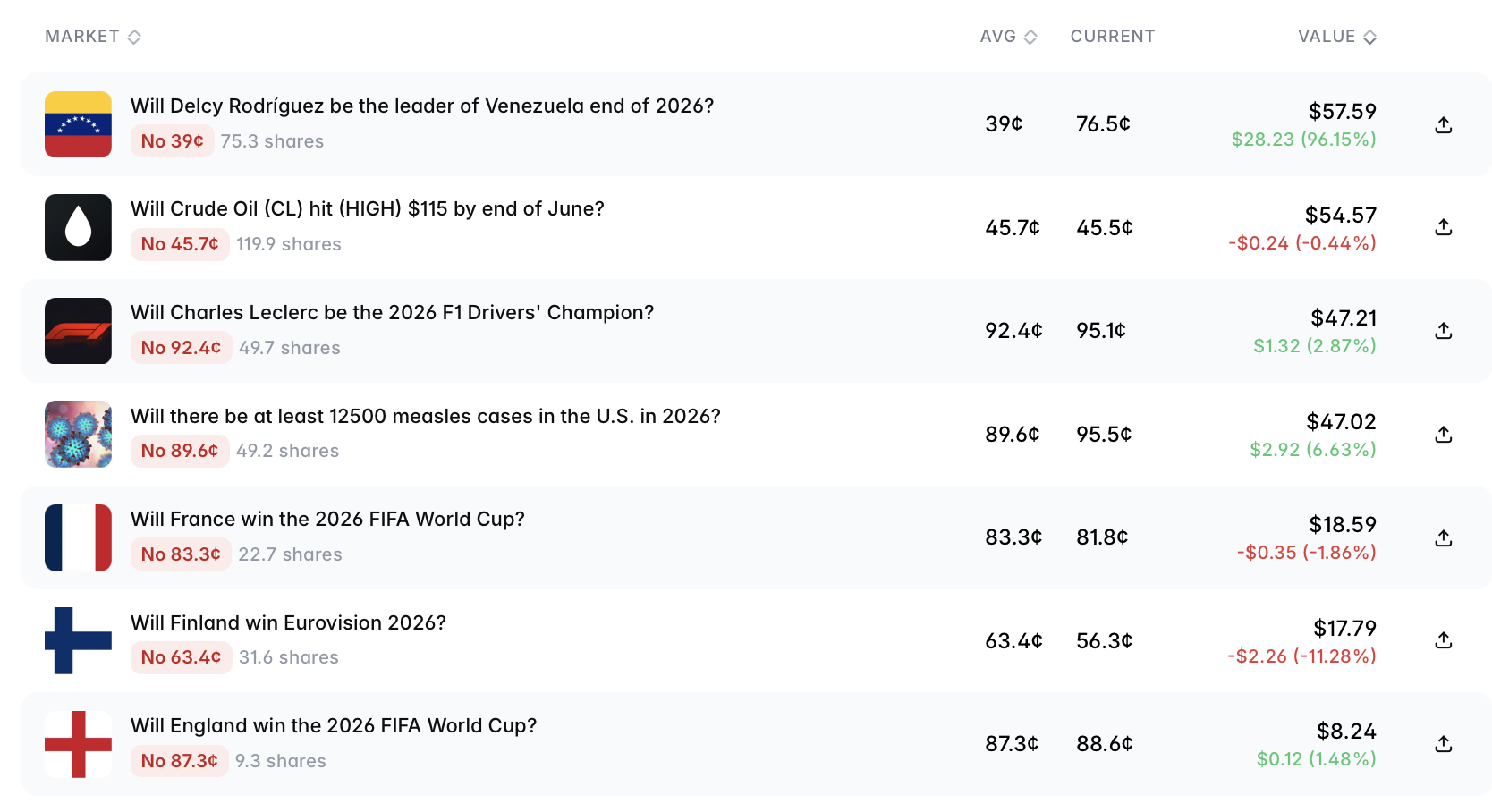}
    \caption{Current open positions: seven NO-side contracts spanning politics, commodities, sports, entertainment, and public health.}
    \label{fig:positions}
  \end{subfigure}
  \hfill
  \begin{subfigure}[t]{0.48\textwidth}
    \centering
    \includegraphics[width=\linewidth]{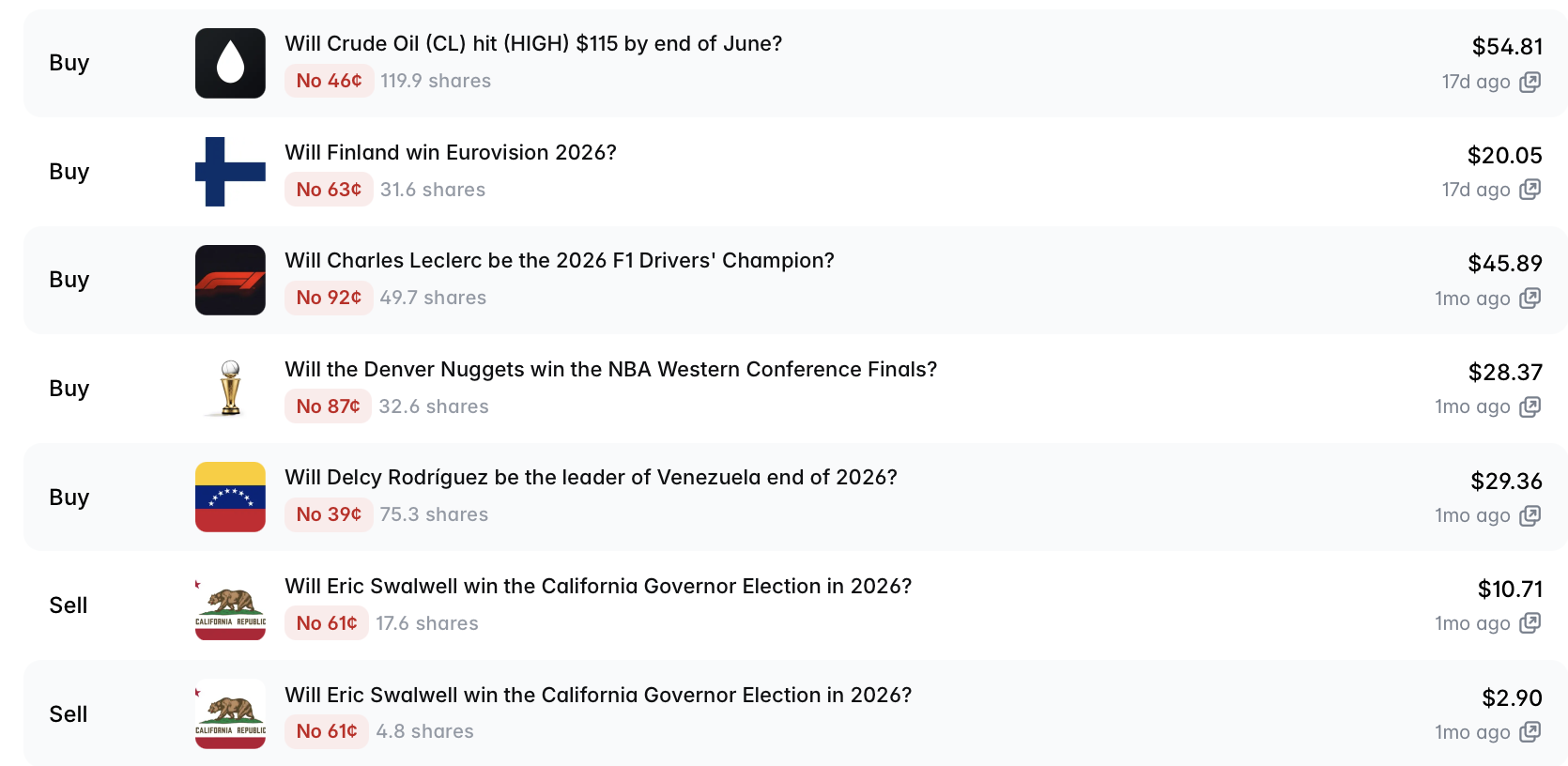}
    \caption{Recent trade history: a mix of NO-side entries (crude oil, Eurovision, F1, NBA, Venezuela).}
    \label{fig:trade-history}
  \end{subfigure}

  \caption{Polymarket dashboard snapshots of Raven-Agent's open positions and recent trade history during live deployment.}
  \label{fig:polymarket-dashboard}
\end{figure*}

Raven-Agent (full) is the only policy with positive ROI and the only one with positive $S_w$ (\cref{tab:main-results}).
On Polymarket, Raven-Agent has also accumulated \$53.92 across 20 predictions in live deployment (\cref{fig:pnl-curve,fig:polymarket-dashboard}).

The Edge-proportional policy shows that sizing alone, without selection, is harmful: concentrating capital on confidently wrong high-edge predictions amplifies losses to $-$55.5\% ROI, far worse than flat sizing ($-$10.7\%).
Raven-Agent (full) (+15.9\%) avoids this by first removing value-destroying forecasts, suggesting that selection and risk filtering are important complements to informed sizing.
Among the fixed-stake policies, each additional trading-layer component improves ROI, and sizing then turns the same selected set from a small loss into a substantial gain.

The Selected Brier decreases along the policy sequence because more aggressive filtering retains better-calibrated rows, not because the forecaster improves (the full-archive Brier is 0.329 for every policy).
Bootstrap confidence intervals and leave-one-out sensitivity are reported in \cref{app:robustness}; an expanded 132-row replay on a fused archive is in \cref{app:fused}.

\section{Conclusion and discussion}

\subsection{Conclusion}

We studied the gap between forecasting and trading on live prediction markets.
We presented Raven-Agent, a system whose trading layer handles selection, sizing, and risk control on top of a swappable forecaster.
On a controlled replay over a fixed forecast archive, Raven-Agent achieved the only positive return and risk-adjusted return among five policies.
The edge-proportional baseline, which sizes by edge without selection, performs substantially worse than flat sizing ($-$55.5\% vs.\ $-$10.7\%), suggesting that selection and risk filtering are important complements to informed sizing.

\subsection{Discussion}

The current results are preliminary.
The sample is modest and the forecaster is held fixed, which isolates the trading layer but leaves the forecaster contribution unmeasured.
A stronger forecaster can be plugged into the same replay infrastructure, and larger archives would strengthen the statistical claims.
Despite these caveats, the replay suggests that explicit selection, sizing, and risk controls materially improve observed outcomes even with the forecaster held constant.

\bibliography{references}
\bibliographystyle{icml2026}

\newpage
\appendix
\onecolumn
\section{Live deployment and code}
\label{app:deployment}

The Raven-Agent deployment discussed in this paper was run on a live Polymarket account.
For transparency, the public Polymarket profile is available at:
\url{https://polymarket.com/profile/0x6664e32f79aee42639f73633e40b5a842b07614e}.
The accompanying code and redacted experiment artifacts are available at:
\url{https://github.com/Alchemist-X/predict-raven}.
The repository contains the agent implementation, replay scripts, and derived tables used in the experiments, with private credentials and non-public runtime paths removed.

\section{Related work}
\label{app:related}

\paragraph{Forecasting benchmarks.}
Evaluation of LLM forecasting has moved from static question answering toward live prediction market settings \citep{karger2024forecastbench,yang2025prophet,zeng2025futurex}.
Prophet Arena \citep{yang2025prophet} reports Brier score, calibration, and market return as three complementary axes.
It shows that frontier language models match a market consensus policy on probability quality but do not break even on return.
PolyBench \citep{cheng2026polybench} couples market snapshots with order book state.
It finds that most evaluated models incur losses under realistic execution, even when their stated probabilities are confident.
Prediction Arena \citep{zhang2026predictionarena} reports the same pattern under live trading.
Our work builds on this evaluation infrastructure, but shifts the unit of analysis from the forecaster to the trading policy.
We hold archived forecasts fixed and vary the surrounding decision and risk modules.

\paragraph{Forecasting agents.}
A complementary line of work pushes the forecasting layer itself.
BLF \citep{murphy2026blf} uses structured linguistic belief states, multi trial aggregation, and hierarchical calibration to substantially improve Brier scores on \citet{karger2024forecastbench}.
These methods are orthogonal to our work.
A stronger forecaster can be inserted into the probability analysis module of Raven-Agent without changing the rest of the system, and the benefits of the two lines compound.
We do not optimize the forecaster in this paper.
We study what should surround it before its outputs control capital.

\paragraph{Risk aware decision making.}
End to end risk aware policy optimization is widely studied in financial reinforcement learning.
Exploration there is sample intensive and tolerates reversible mistakes.
On live prediction markets, every exploratory action consumes real bankroll on a market that resolves within days, which makes large scale policy training infeasible.
We therefore implement risk constraints as deterministic services outside the language model prompt.
This bounds the worst case loss per decision and produces auditable rejection reasons that learned policies cannot easily match.
Recent concurrent work provides direct empirical evidence for this design choice.
\citet{zhang2026predictionarena} deployed six frontier language models as end-to-end prediction market agents on Kalshi, each trading with a \$10{,}000 bankroll over 57 days.
All six lost money, with returns ranging from $-$16\% to $-$31\%.
The only risk controls that held were hard-coded constraints (a 15\% per-market concentration cap and solvency checks); prompt-level risk guidance was frequently ignored.
\citet{jang2025losingwinner} show that even with a deterministic execution rule, optimizing a proxy forecasting objective (classification accuracy via RLVR) can improve accuracy while worsening trading return, with their best-accuracy agent achieving $-$14.8\% return.
Together, these findings motivate separating the trading layer from the language model and enforcing risk constraints outside the prompt.

\paragraph{Comparison with Prediction Arena.}
Prediction Arena \citep{zhang2026predictionarena} and Raven-Agent differ in a key architectural choice.
Prediction Arena gives each language model end-to-end control: the model receives market data and portfolio state, reasons freely with tool access, and directly issues trade orders with no external sizing or risk module.
Raven-Agent separates the forecasting module from the trading layer, so selection, sizing, and risk enforcement are deterministic components that the language model cannot override.
The two setups are therefore not directly comparable on the same leaderboard: Prediction Arena evaluates the LLM as a complete trading agent, while our replay protocol evaluates the trading layer with the forecaster held fixed.
A controlled comparison would require running Prediction Arena agents on our archived decision set, substituting their end-to-end decisions for our trading layer while keeping the same market snapshots and price data.
We leave this to future work.

\section{Archive coverage}
\label{app:archive}

The replay archive records the intermediate outputs produced by Raven-Agent during deployment.
It starts from raw market scans, then applies filtering, candidate persistence, recommendation generation, and final trading decisions.
The table below reports how many artifacts are available at each stage of this pipeline.
These counts are intended to make the replay scope clear: many markets are scanned, a smaller set is kept as candidates, and only a subset becomes executable trading decisions.

\paragraph{Replay procedure.}
In each archived run, the agent first queried Polymarket markets and filtered them by liquidity, resolution clarity, usable prices, and event overlap.
The candidate set included markets such as crude-oil threshold contracts, political resignation markets, sports futures, and other high-liquidity event markets.
For example, one archived run scanned 1{,}544 Polymarket markets, retained 92 after filtering, and selected 12 candidates for deeper analysis.
The final recommendations in that run focused on the No side of crude oil reaching \$100 by the end of March, Bitcoin dipping to \$65{,}000 in March, and Netanyahu leaving office by June 30.

For each retained candidate, Raven-Agent estimated the probability that the selected contract would pay out.
It compared this probability with the Polymarket price, computed the edge, ranked opportunities, and proposed a stake.
When the live system decided to trade, it submitted real orders through Polymarket's CLOB API rather than simulating execution inside the model.
The replay uses the archived forecasts, prices, order metadata, and execution records from these live runs.

\begin{table}[h]
  \centering
  \caption{Archive coverage used by the replay experiments. Counts are grouped by the replay pipeline stage.}
  \label{tab:coverage}
  \small
  \begin{tabular}{@{}llr@{}}
    \toprule
    \textbf{Stage} & \textbf{Quantity} & \textbf{Count} \\
    \midrule
    Raw scan & Pulse JSON files & 65 \\
    Raw scan & Fetched market observations & 264{,}859 \\
    Filtering & Filtered market observations & 25{,}231 \\
    Candidate set & Persisted candidate rows & 844 \\
    Candidate set & Unique persisted candidate markets & 222 \\
    Recommendation & Recommendation files & 45 \\
    Decision & Decision rows & 250 \\
    Decision & Open-decision rows & 59 \\
    Decision & Executable-plan rows & 44 \\
    Execution & Matched observed fills & 18 \\
    \midrule
    Expansion & Earlier selected Full Raven rows & 90 \\
    Expansion & Combined priced rows & 132 \\
    \bottomrule
  \end{tabular}
\end{table}

\section{Robustness analysis}
\label{app:robustness}

\paragraph{Bootstrap confidence intervals.}
We resample trades with replacement 10{,}000 times and report 95\% percentile intervals for ROI and $S_w$ (\cref{tab:bootstrap}).
Raven-Agent (full) is the only policy whose ROI confidence interval lies entirely above zero.
Edge-proportional's interval lies entirely below zero, confirming that the negative result is robust and not an artifact of a few extreme trades.

\begin{table}[h]
  \centering
  \caption{Bootstrap 95\% confidence intervals (10{,}000 resamples).}
  \label{tab:bootstrap}
  \small
  \begin{tabular}{@{}lcc@{}}
    \toprule
    \textbf{Policy} & \textbf{ROI 95\% CI} & \textbf{$S_w$ 95\% CI} \\
    \midrule
    Forecast-only & [$-$24.7\%, +3.2\%] & [$-$0.46, +0.06] \\
    Edge-proportional & [$-$71.8\%, $-$29.2\%] & [$-$1.42, $-$0.44] \\
    Edge-filter & [$-$25.1\%, +6.8\%] & [$-$0.46, +0.12] \\
    Raven (fixed) & [$-$20.6\%, +11.5\%] & [$-$0.41, +0.22] \\
    Raven (full) & [+3.8\%, +30.0\%] & [+0.11, +0.76] \\
    \bottomrule
  \end{tabular}
\end{table}

\paragraph{Leave-one-out sensitivity.}
We compute ROI after dropping each trade in turn (\cref{tab:loo}).
Raven-Agent (full) ROI stays positive across all deletions, ranging from +12.7\% to +19.0\%, indicating that no single trade drives the result.

\begin{table}[h]
  \centering
  \caption{Leave-one-out ROI sensitivity.}
  \label{tab:loo}
  \small
  \begin{tabular}{@{}lcc@{}}
    \toprule
    \textbf{Policy} & \textbf{ROI range} & \textbf{Std} \\
    \midrule
    Forecast-only & [$-$13.6\%, $-$9.1\%] & 1.0\% \\
    Edge-proportional & [$-$58.0\%, $-$51.6\%] & 1.5\% \\
    Edge-filter & [$-$12.8\%, $-$7.4\%] & 1.2\% \\
    Raven (fixed) & [$-$8.5\%, $-$2.4\%] & 1.3\% \\
    Raven (full) & [+12.7\%, +19.0\%] & 1.0\% \\
    \bottomrule
  \end{tabular}
\end{table}

\section{Expanded replay on a fused archive}
\label{app:fused}

To probe robustness on a larger sample, we extend the replay to an earlier Raven-Agent deployment that produced 90 additional selected recommendations.
The combined 132-row set yields a positive ROI of +5.9\% under fixed \$10 stakes (\cref{tab:fused-results}), confirming the main-archive direction on a larger sample.
Bootstrap 95\% CI for the fused set is [$-$4.5\%, +16.1\%] for ROI and [$-$0.07, +0.27] for $S_w$.
The earlier archive contains only selected trades (the rejected diagnostic cannot be extended), and the archived sizing column is omitted because the earlier deployment used a much larger bankroll (${\sim}$\$100k vs.\ ${\sim}$\$1k), making absolute stake and PnL incomparable.

\begin{table}[h]
  \centering
  \caption{Fused Raven-Agent selected-trade replay (fixed \$10 stake).}
  \label{tab:fused-results}
  \small
  \begin{tabular}{@{}lrrrrrr@{}}
    \toprule
    \textbf{Dataset} & \textbf{Priced} & \textbf{Stake} & \textbf{PnL} & \textbf{ROI} & \textbf{$S_w$} & \textbf{Brier} \\
    \midrule
    Main archive & 42 & \$420 & $-$\$20 & $-$4.7\% & $-$0.09 & 0.258 \\
    Earlier archive & 90 & \$900 & +\$98 & +10.9\% & +0.17 & 0.188 \\
    Combined & 132 & \$1{,}320 & +\$78 & +5.9\% & +0.10 & 0.204 \\
    \bottomrule
  \end{tabular}
\end{table}

\section{Example archived agent traces}
\label{app:archived-traces}

\begin{figure*}[t]
  \centering
  \includegraphics[width=0.95\textwidth]{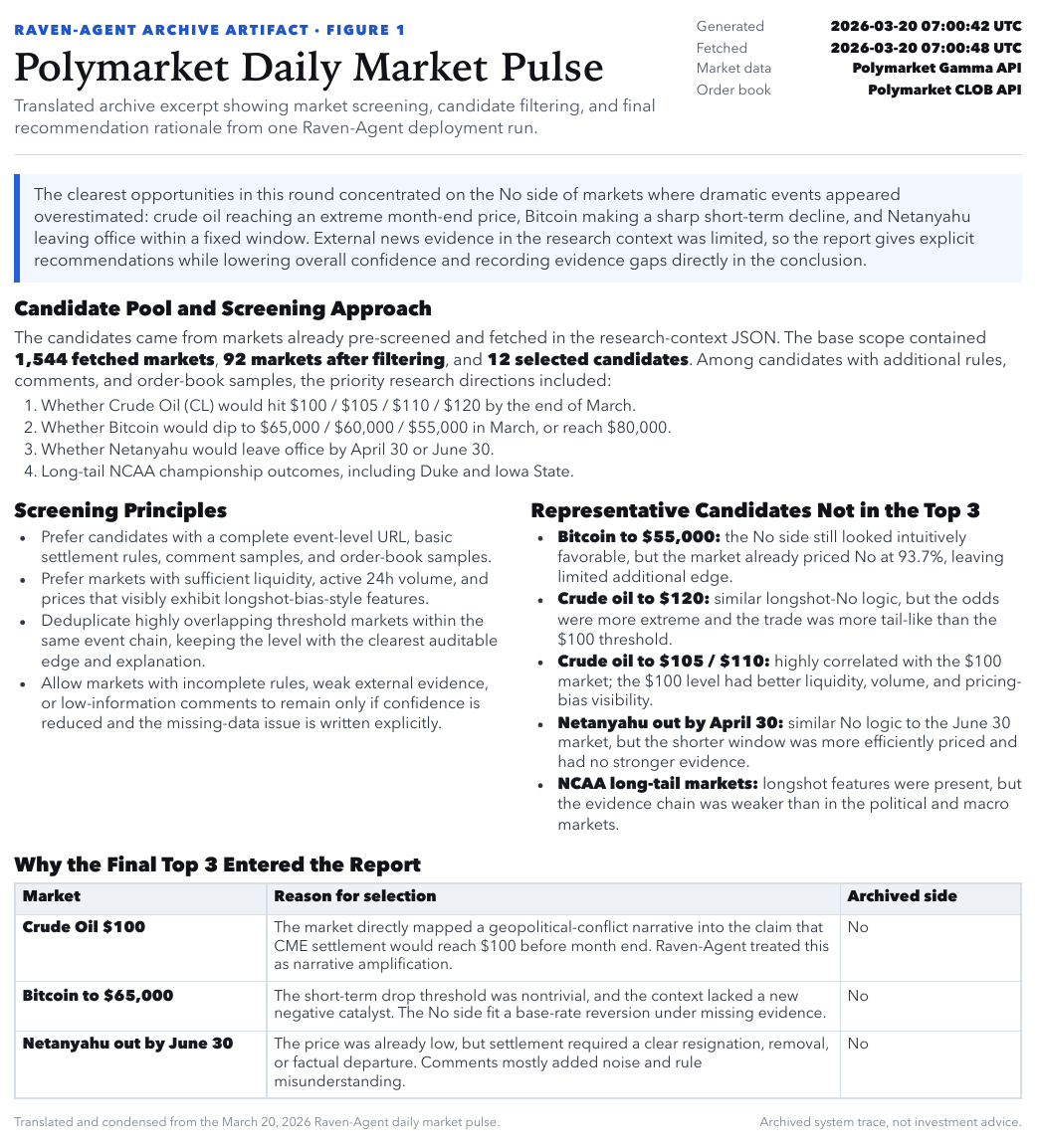}
  \caption{
  Example Raven-Agent market screening trace.
  The artifact records the daily scan, filtering rules, representative rejected candidates, and final selected markets from one archived market pulse.
  }
  \label{fig:archive-screening-trace}
\end{figure*}

\begin{figure*}[t]
  \centering
  \includegraphics[width=0.95\textwidth]{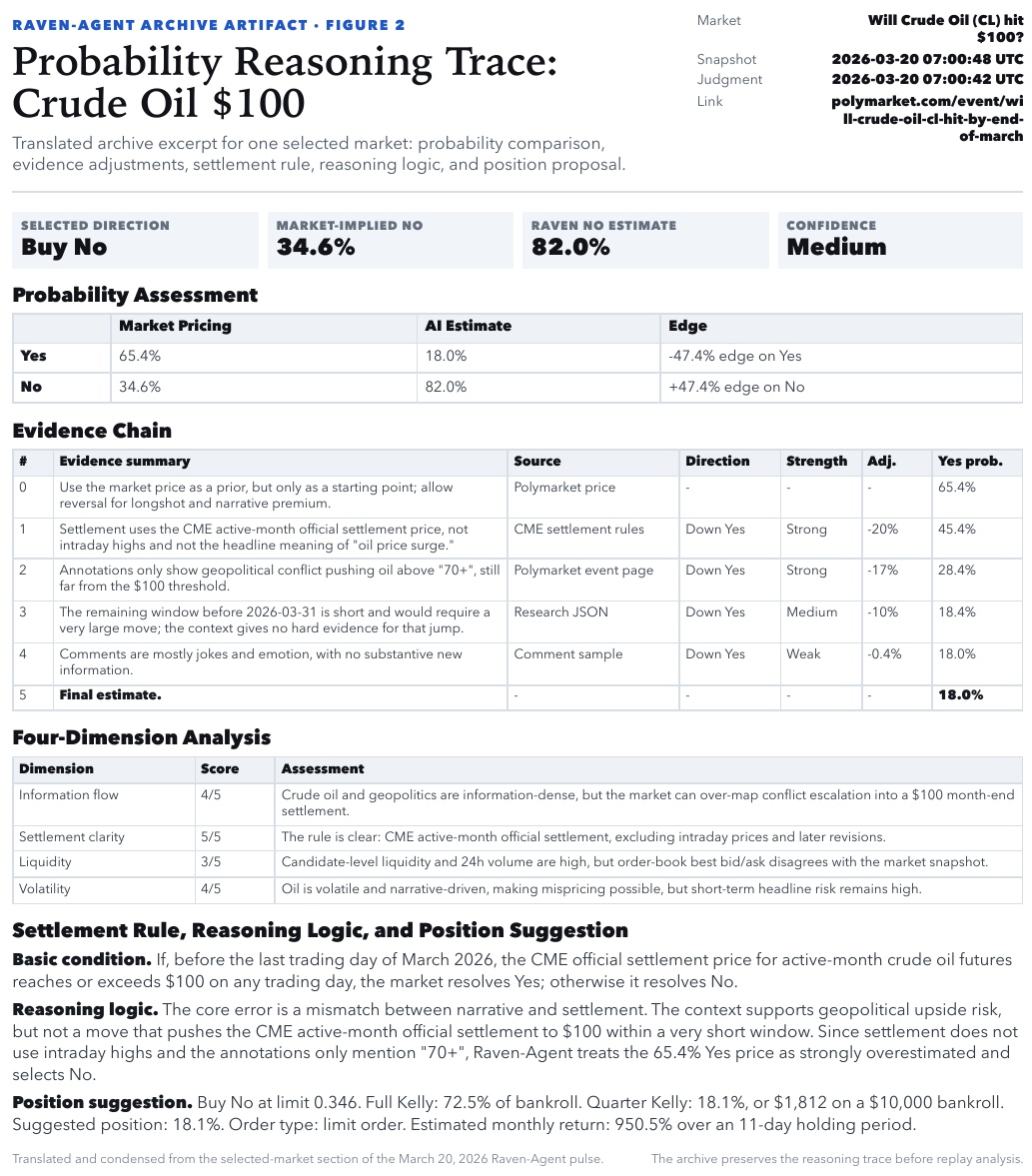}
  \caption{
  Example Raven-Agent probability reasoning trace for a selected crude oil market.
  The artifact records the market-implied probability, Raven-Agent probability estimate, evidence adjustments, settlement-rule checks, and proposed action.
  }
  \label{fig:archive-probability-trace}
\end{figure*}

Raven-Agent stores intermediate artifacts during deployment, not only final trades.
These artifacts are useful for auditing the belief-to-trade pipeline because they record which markets were scanned, which candidates were filtered, what probability estimates were produced, and why a selected market was considered actionable.
\Cref{fig:archive-screening-trace} shows an example daily market pulse after translation and condensation.
The trace begins with a broad market scan, narrows the pool through filtering and candidate selection, and records the rationale for the final Top 3 recommendations.
\Cref{fig:archive-probability-trace} shows the corresponding probability reasoning trace for one selected market.
It preserves the market-implied probability, the agent probability, evidence adjustments, settlement-rule checks, confidence notes, and the resulting trading proposal.

These figures are not additional quantitative results.
They are included to make the archived decision process concrete and to show that the replay is based on structured records left by the deployed agent rather than on reconstructed explanations after the fact.
Private credentials and non-public runtime paths are omitted.

\end{document}